\pdfoutput=1

\documentclass[11pt]{article}

\usepackage{acl}

\usepackage{times}
\usepackage{latexsym}
\usepackage{arydshln}
\usepackage{amsmath,amssymb}
\usepackage{algorithm, algorithmic}

\usepackage[subrefformat=parens]{subcaption}
\usepackage{svg}
\usepackage{bm}
\usepackage{url}
\usepackage{graphicx}
\usepackage{CJKutf8}

\usepackage[T1]{fontenc}

\usepackage[utf8]{inputenc}

\usepackage{microtype}

\def\eqref#1{equation~\ref{#1}}

\def\1{\bm{1}}

\def\vb{{\bm{b}}}
\def\vc{{\bm{c}}}

\def\vs{{\bm{s}}}
\def\vt{{\bm{t}}}

\def\vw{{\bm{w}}}
\def\vx{{\bm{x}}}

\def\mD{{\bm{D}}}
\def\mE{{\bm{E}}}

\def\mQ{{\bm{Q}}}

\def\mT{{\bm{T}}}

\def\mW{{\bm{W}}}

\DeclareMathAlphabet{\mathsfit}{\encodingdefault}{\sfdefault}{m}{sl}
\SetMathAlphabet{\mathsfit}{bold}{\encodingdefault}{\sfdefault}{bx}{n}

\title{Unsupervised Domain Adaptation for Sparse Retrieval \\ by Filling Vocabulary and Word Frequency Gaps}

\author{Hiroki Iida \and Naoaki Okazaki \\
 Department of Computer Science, School of Computing, Tokyo Institute of Technology \\
   {\texttt \{hiroki.iida@nlp.c., okazaki@c.\}titech.ac.jp} \\}

\begin{document}
\maketitle
\begin{abstract}
IR models using a pretrained language model significantly outperform lexical approaches like BM25. In particular, SPLADE, which encodes texts to sparse vectors, is an effective model for practical use because it shows robustness to out-of-domain datasets. However, SPLADE still struggles with exact matching of low-frequency words in training data. In addition, domain shifts in vocabulary and word frequencies deteriorate the IR performance of SPLADE. Because supervision data are scarce in the target domain, addressing the domain shifts without supervision data is necessary.
This paper proposes an unsupervised domain adaptation method by filling vocabulary and word-frequency gaps. First, we expand a vocabulary and execute continual pretraining with a masked language model on a corpus of the target domain. Then, we multiply SPLADE-encoded sparse vectors by inverse document frequency weights to consider the importance of documents with low-frequency words.  
We conducted experiments using our method on datasets with a large vocabulary gap from a source domain. We show that our method outperforms the present state-of-the-art domain adaptation method. In addition, our method achieves state-of-the-art results, combined with BM25. 

\end{abstract}

\section{Introduction}
Information retrieval (IR) systems are widely used nowadays. Most of them are based on lexical approaches like BM25~\citep{Robertson1994-ci}. Because lexical approaches are based on bag-of-words (BoW), they suffer from \textit{vocabulary mismatch}, where different words express the same notion. Recently, IR models with a pre-trained masked language model (MLM), such as BERT~\citep{devlin-etal-2019-bert} have overcome this problem and outperformed BM25~\citep{Nogueira2019-hc, Karpukhin2020-dm, xiong2021approximate, Formal2021-hj}. 

In particular, SPLADE~\citep{Formal2021-hj} is an effective model for practical use. SPLADE addresses vocabulary mismatch by expanding queries and documents through an MLM. Concretely, SPLADE encodes texts to sparse vectors using the logits of the MLM for each token of the texts. As a result, each element of these vectors corresponds to a word in the vocabulary of the MLM. In addition, the nonzero elements other than tokens appearing in the texts can be considered as query and document expansion. Because the encoded vectors are sparse, SPLADE can realize a fast search by utilizing inverted indexes and outperforms BM25, even when SPLADE is applied to out-of-domain datasets from a source domain of training data. 

However, SPLADE still struggles with the exact matching of low-frequency words in the training data~\citep{Formal2021-zs}. This problem is amplified for out-of-domain datasets. In addition, \citet{Thakur2021-rt} discussed that large domain shifts in vocabulary and word frequencies deteriorate the performance of vector-based IR models. 
Furthermore, preparing massive supervision data for every dataset is impractical due to annotation costs.
Thus, a method to address the domain shifts without supervision data is necessary.

Unsupervised domain adaptation (UDA) is an approach to overcome domain shift without supervision data. However, as discussed in Section~\ref{analysis-rsj-weight}, generated pseudo labeling (GPL) ~\citep{noauthor_undated-vv}, a state-of-the-art UDA method using generated queries, cannot solve the problem of low-frequency words on some datasets.

\begin{figure}[t]
  \centering
  \includegraphics[keepaspectratio, scale=0.42]{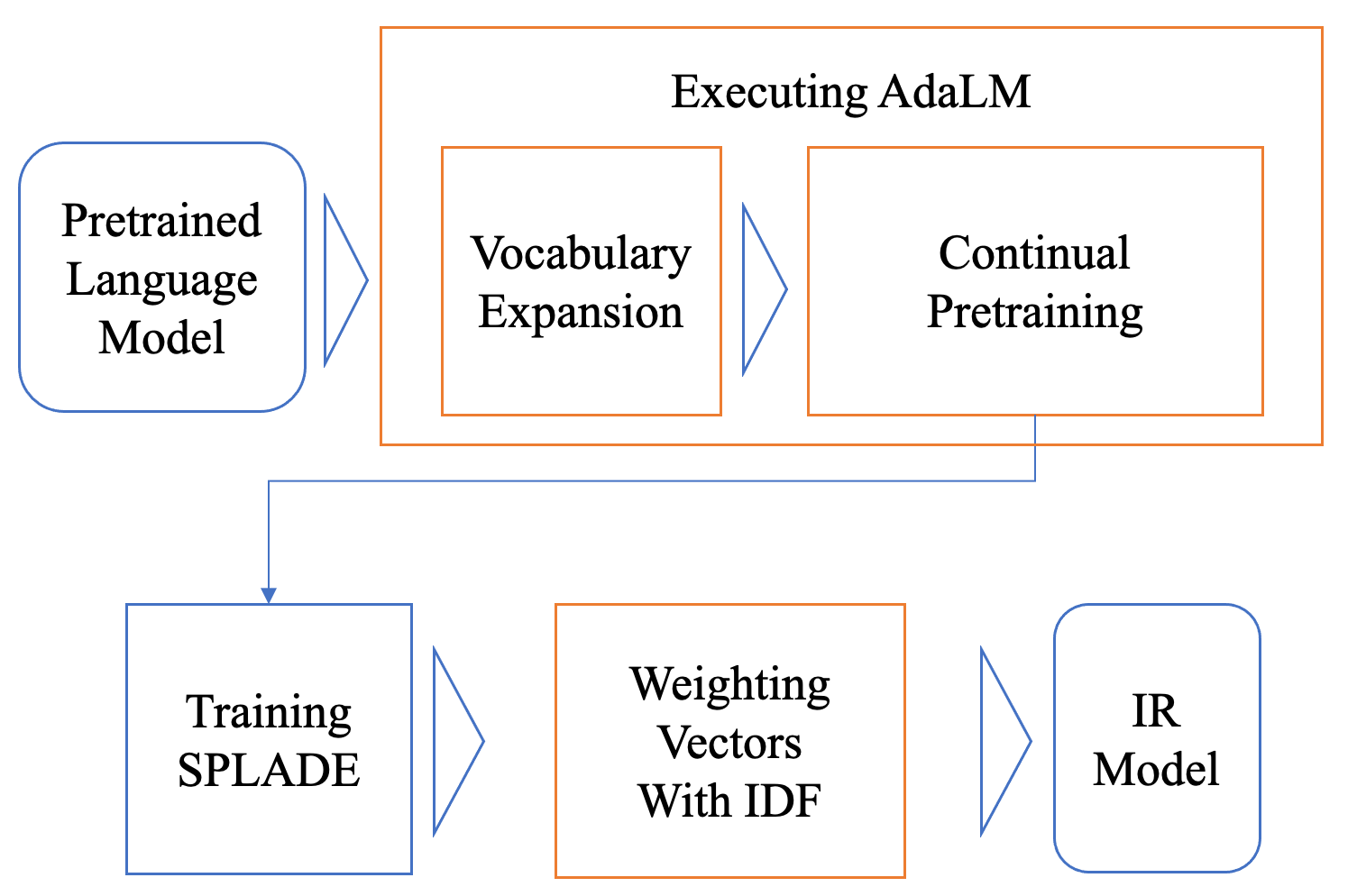}
  \caption{Outline of our method. Orange boxes indicate our proposal.}
  \label{fig-outline}
\end{figure}

In this paper, we propose a UDA method that fills the vocabulary and word-frequency gap between the source and target domains. Specifically,  we use AdaLM~\citep{yao-etal-2021-adapt}, which is a domain adaptation method for an MLM through vocabulary expansion~\citep{wang-etal-2019-improving, hong-etal-2021-avocado} and continual pre-training~\citep{gururangan-etal-2020-dont} on a domain-specific corpus. 
We expect AdaLM to realize more accurate query and document expansions in the target domain. Furthermore, because SPLADE struggles with exact matching of low-frequency words in the training data, we weight such words by multiplying each element of the SPLADE-encoded sparse vectors by inverse document frequency (IDF) weights. We call our method Combination of AdaLM and IDF (CAI).

We apply CAI to SPLADE and conducted experiments with it on five IR datasets from bio-medical and science domains in the BEIR benchmark~\citep{Thakur2021-rt}. We used these datasets because the five datasets have the largest vocabulary gap in BEIR from MS MARCO~\citep{DBLP:conf/nips/NguyenRSGTMD16}, the source dataset.
The experimental results show that SPLADE with CAI outperforms SPLADE with GPL and achieves state-the-art results on average across all datasets by adding scores of BM25.

Finally, to confirm whether CAI can address the problem of exact matching words of low-frequency words in training data, we analyzed the weights of exact matching words, following the approach of~\citet{Formal2021-zs}. Our analysis confirms that SPLADE with CAI addresses the problem of the exact matching, whereas SPLADE with GPL cannot.

Our contributions can be summarized as follow:
\begin{itemize}
    \item We present an unsupervised domain adaptation method, filling vocabulary and word-frequency gaps between the source and target domains. Furthermore, we show that our method performs well in sparse retrieval.
    \item We confirm that CAI outperforms GPL, the state-of-the-art domain adaptation method for IR, on datasets with large domain shifts from a source dataset.
    \item Our analysis shows that a factor in the success of CAI is addressing the problem of exact matching of low-frequency words. 
\end{itemize}

\section{Related Works}
~\citet{Thakur2021-rt} showed that vector-based IR models based on a pretrained MLM deteriorate when applied to out-of-distribution datasets. They discussed that one of the causes of the deterioration of the IR performance was a large domain shift in vocabulary and word frequencies. ~\citet{Formal2021-zs} also found that IR models based on an MLM struggled with exact matching of low-frequency words in training data. This problem also leads to performance deterioration of MLM-based IR models on out-of-distribution datasets. ~\citet{MacAvaney2020-xd} showed that a domain-specific MLM performed better than an MLM trained on a corpus of a general domain. However, no previous works showed that addressing vocabulary and word-frequency gaps can solve the problem of deterioration of IR performance for vector-based IR models.

Unsupervised domain adaptation (UDA) is a promising approach to solve the degradation due to domain shift without supervision data. MoDIR~\citep{xin-etal-2022-zero} adopts domain adversarial loss~\citep{Ganin2017-xx} to allow a dense retrieval model to learn domain-invariant representations. Other approaches utilize generated queries. GenQ~\cite{ma-etal-2021-zero} generates queries from a document in an IR corpus with a generative model and then considers the pairs of generated queries and a document as relevant pairs. In addition to GenQ, GPL~\citep{noauthor_undated-vv} uses documents retrieved by an IR model against a generated query as negative examples and adopts Margin-MSE loss~\citep{Hofstatter2020-tz}, which discerns how negative the retrieved documents are. GPL outperforms MoDIR, continual pretraining~\citep{gururangan-etal-2020-dont}, and UDALM~\citep{karouzos-etal-2021-udalm}. However, these approaches target dense representations, and their effect on sparse representation is unclear. We present a more effective UDA method, especially for sparse representations.

\section{Method}
This paper proposes the UDA method to tackle the domain shifts in vocabulary and word frequency. An outline of our method is illustrated in Figure~\ref{fig-outline}. Our method consists of three parts: (1) executing AdaLM for domain adaptation of an MLM, (2) training SPLADE with supervised data, and (3) weighting sparse vectors encoded by SPLADE with IDF when searching. Our proposal parts are (1) and (3). We first introduce SPLADE as preliminary.


\subsection{SPLADE (Preliminary)}

 SPLADE~\citep{Formal2021-hj} is a supervised IR model. The model encodes queries and documents to sparse vectors using logits of an MLM and calculates relevance scores by dot products of sparse vectors of the queries and documents. 

Let $\mathcal{V}$ denote the vocabulary of an MLM. We represent a text $T$ as a sequence of $n+2$ tokens, $T = (t_0, t_1, t_2, \dots, t_n, t_{n+1}) \in \mathcal{V}^{n+2}$, where $t_0$ represents the CLS token and $t_{n+1}$ represents the SEP token. Each token is encoded to a $d$-dimensional vector,  $\vt_i \in \mathbb{R}^{d}$, using the MLM. We express the sequence of $n+2$ encoded tokens as $\mT = (\vt_0, \vt_1, \vt_2, \dots, \vt_n, \vt_{n+1})$.

We express the process of SPLADE encoding $T$ to a sparse vector $\vs \in \mathbb{R}^{|\mathcal{V}|}$ as $\text{SPLADE}(T)$. Formally, we say 

\begin{equation}
    \vs = \text{SPLADE}(T).
\end{equation}

Now, we explain the encoding process. First, the text $T$ is encoded to $\mT$. Then, SPLADE converts $\vt_i \in \mT$ to a sparse vector $\vs_{i} \in \mathbb{R}^{|\mathcal{V}|}$ through the MLM layer. The formal expression is

\begin{equation}
\label{eq:decode}
    \vs_{i} = \mE f(\mW\vt_i + \vb) + \vc.
\end{equation}

Here, $\mE \in \mathbb{R}^{\mathcal{|V|} \times d}$ is the embedding layer of the MLM. Then, $\vc \in \mathbb{R}^{|\mathcal{V}|}$ is a bias term of the embedding layer. $\mW \in \mathbb{R}^{l \times l}$ is a linear layer, and $\vb \in \mathbb{R}^{l}$ is a bias term of the linear layer. $f()$ is an activation function with LayerNorm. 

Then, we obtain a sparse vector $\vs$ by max-pooling with log-saturation effect:

\begin{equation}
\label{eq:max-pool}
    \vs = \max_{0 \le i \le n+1} \log(\textbf{1}+\text{ReLU}(\vs_{i})).
\end{equation}
Here, the sparse vector $\vs$ is likely to have nonzero elements other than $t_i \in T$. In this sense, SPLADE can be considered as a method using query and document expansion.

SPLADE infers a relevance score by inner product between sparse vectors of a query and document. We denote a tokenized query as $Q \in \mathcal{V}^{l}$ and a tokenized document as $D \in \mathcal{V}^{m}$. $l$ and $m$ are the lengths of $Q$ and $D$, respectively. The relevance score of SPLADE, $S_{\text{SPL}}(Q, D)$, is formally
\begin{equation}
    S_{\text{SPL}}(Q, D) = \text{SPLADE}(Q)^{\top} \text{SPLADE}(D).
\end{equation}

Practically, reducing computational cost is another important point, especially when searching. ~\citet{Formal2021-hj} replaced $\text{SPLADE}(Q)$ with a bag-of-words (BoW) representation of a query. 
\citet{Formal2021-hj} called this scoring \textbf{SPLADE-Doc}. This case gives no query expansion. Formally, the score of SPLADE-Doc, $S_{\text{SPL-D}}(Q, D)$, is 
\begin{gather}
    S_{\text{SPL-D}}(Q, D) = \sum_{t \in Q} \text{SPLADE}(D)_t.
\end{gather}
Here, $\text{SPLADE}(D)_t$ is the element $t$ of a sparse vector of $D$. Note that a token $t \in Q$ can indicate the element of the sparse vector of $D$.

To learn sparse representations, SPLADE adopts the FLOPS regularizer~\cite{Paria2020Minimizing}. We give the formal expression of the FLOPS regularizer in Appendix~\ref{sec:flops}. 


\subsection{Combination of AdaLM and IDF}
\label{sec:CAI}

This section explains our proposed CAI method more precisely.

\subsubsection{Executing AdaLM}
CAI is a method addressing the vocabulary and word-frequency gap between datasets without supervision data. We execute AdaLM before training SPLADE to fill this gap. AdaLM~\citep{yao-etal-2021-adapt} is a UDA method for an MLM. It comprises vocabulary expansion and continual pretraining using the corpus of the target domain. 

We use AdaLM based on two assumptions. One is that we can consider that SPLADE expands queries and documents because the sparse vector encoded by SPLADE has non-zero elements corresponding to tokens that do not appear in a query or document. Thus, continual pretraining should allow SPLADE to expand queries and documents more accurately. In addition, vocabulary expansion should amplify the effect of continual pretraining.
The other is that~\citet{jang-etal-2021-ultra} showed that the larger the dimension of sparse vectors, the better sparse retrieval performed in MRR@10 in the source domain. Vocabulary expansion means increasing dimensions of sparse vectors for SPLADE. Thus, we expect that vocabulary expansion should improve IR performance even on out-of-domain datasets.

In the last part of this subsection, we explain details of how to execute AdaLM.
\paragraph{Vocabulary Expansion}
\label{subsec-vocab-expand}
AdaLM first expands the vocabulary of an MLM for more effective continual pretraining. To expand the vocabulary, AdaLM first builds a domain-specific tokenizer with WordPiece~\cite{Schuster2012JapaneseAK} at a target vocabulary size. Then, AdaLM adds new words obtained by the built tokenizer to the original tokenizer. The addition starts from the most frequent words and stops when the vocabulary size of the tokenizer reaches the target vocabulary size. We exclude tokens composed of only numbers and marks (e.g., !,?,",[,]) because these tokens are considered as noise. We repeat this procedure, increasing the target vocabulary by 3k. Finally, we stop this increment when the vocabulary size of the tokenizer cannot reach the target vocabulary size. We summarize this procedure in Algorithm~\ref{algo-expand-voc}. 

After adding words, AdaLM initializes the embeddings of these words. To obtain the embeddings, AdaLM tokenizes the added words to subwords by the original tokenizer, takes the average of embeddings of subwords, and then sets the averaged embeddings as initial vectors of newly added words.

\begin{algorithm}[!t]
\caption{Procedure for vocabulary expansion}
\label{algo-expand-voc}
\begin{algorithmic}[1]
\STATE \textbf{INPUT}: Original vocabulary $\mathcal{V}_0$, a domain corpus $C$, incremental vocabulary size $\Delta V$
\STATE \textbf{OUTPUT}: $\mathcal{V}_{\text{final}}$
\STATE Set iterating index $i=0$
\REPEAT
  \STATE $i = i + 1$
  \STATE $\mathcal{V}_i = \mathcal{V}_{i-1}$
  \STATE Set target vocabulary size $V_i = |\mathcal{V}_0| + i * \Delta V$
  \STATE Build WordPiece tokenizer $\mathcal{T}_i$ at the vocabulary size of $V_i$ on $C$. 
  \STATE Get vocabulary $\mathcal{\acute{V}}_i$ from $\mathcal{T}_i$\
  \STATE Tokenize $C$ by $\mathcal{T}_i$ and count tokens
  \STATE Sort $\mathcal{\acute{V}}_i$ by frequency
  \STATE Set new vocabulary $\mathcal{V}_i$ by adding words to $\mathcal{V}_0$ from frequent words until $|\mathcal{\acute{V}}_i| < V_i$ except for duplicate words and words consisting of only number of mark.
\UNTIL{$ |\mathcal{V}_i| - |\mathcal{V}_{i-1}| < \Delta V$}
\RETURN $\mathcal{V}_{\text{final}} = \mathcal{V}_i$
\end{algorithmic}
\end{algorithm}

\paragraph{Continual Pretraining} Continual pretraining~\citep{gururangan-etal-2020-dont} is also a UDA method for an MLM. This method is straightforward; it further trains an MLM on a domain-specific corpus. Following BERT, we randomly mask 15\% of tokens with a special token like [MASK] and let the model predict the original token.

\subsubsection{Weighting Sparse Vectors with IDF}
After training SPLADE, we multiply the SPLADE-encoded sparse vectors by IDF weights. ~\citet{Formal2021-zs} noted that SPLADE struggles with the exact matching of low-frequency words in the training data. In addition, the problem is amplified on out-of-domain datasets. Thus, we expect sparse vectors weighted with IDF to match the low-frequency words.

Now, we denote the number of documents in a target dataset as $N$ and documents including token $t$ as $N_t$. We express the IDF weight vector $\vw^{\text{IDF}} \in \mathbb{R}^{|\mathcal{V}|}$ by the following equation:

\begin{equation}
    \\ \vw^{\text{IDF}}_t = \\
    \begin{cases}
      \log \frac{N}{N_t} & {\text{if} \, N_t \neq 0}\\
      1 & {\text{otherwise}}
    \end{cases}.
\end{equation}
When $N_t=0$, we set the weight as $1$ so that the weight inferred by SPLADE does not change.

We can express the weighted sparse vector $\hat{\vs} \in \mathbb{R}^{\mathcal{V}}$ by the following equation:

\begin{equation}
\label{eq:weight}
    \hat{\vs} = \vw^{\text{IDF}} \odot \vs.
\end{equation}
where $\odot$ denotes the Hadamard product. Note that we apply the weighting only for document vectors.

\subsection{Combination with Lexical Approach}
 Finally, we discuss the combination of our method with the lexical approach, which is an approach to enhance IR performance further. Previous works showed that lexical approaches and IR models based on an MLM are complementary~\citep{Luan2021-mx, Gao2021-mr}. In addition, several works~\citep{ma-etal-2021-zero, xu-etal-2022-laprador, Formal2022-zx} showed that simply adding or multiplying the scores of the lexical approach and an IR model based on an MLM improved the IR performance. Following these works, we also experimented with the adding case using BM25 for the lexical approach. We refer to this approach as \textbf{Hybrid}. Now, we denote a score of BM25 between a query $Q$ and a document $D$ as $S_{\text{BM25}}(Q, D)$. Formally, for both $S_{\text{SPL}}(Q, D)$ and $S_{\text{SPL-D}}(Q, D)$, the scores of Hybrid, $S_{\text{H-SPL}}(Q, D)$ and $S_{\text{H-SPL-D}}(Q, D)$, are

\begin{gather}
     S_{\text{H-SPL}}(Q, D) = S_{\text{BM25}}(Q, D) + S_{\text{SPL}}(Q, D), \\
    S_{\text{H-SPL-D}}(Q, D) = S_{\text{BM25}}(Q, D) + S_{\text{SPL-D}}(Q, D).
\end{gather}

\section{Experimental Setup}
\label{sec:experimental-seup}
We confirm the effectiveness of our proposed CAI through experimental results. First, we introduce baselines. They show us how effective CAI is. Second, we explain IR datasets and domain corpora. The last subsection gives details of the implementation.\footnote{Our code is available at \url{https://github.com/meshidenn/CAI.git}.}

\subsection{Baselines}
To measure the effectiveness of our approach, we compared it with other IR models. First, we chose dense retrieval~\citep{Karpukhin2020-dm, xiong2021approximate}, Cross Encoder~\citep{Nogueira2019-hc, MacAvaney2019-wr} and LaPraDoR~\citep{xu-etal-2022-laprador}.

\textbf{Dense retrieval} converts queries and documents into dense vectors and calculates relevance scores by the inner product or cosine similarity of dense vectors. Following~\citet{reimers-gurevych-2019-sentence}, we used average pooling to obtain dense vectors and cosine similarity for calculating relevance scores.

\textbf{Cross Encoder}\footnote{The actual model is cross-encoder/ms-marco-MiniLM-L-6-v2.}  lets an MLM infer relevance scores by inputting texts composed from concatenations of queries and documents. This method achieved the best performance in a study by ~\citet{Thakur2021-rt}. We explain Cross Encoder formally in Appendix~\ref{sec:margin-mse}.

\textbf{LaPraDoR} adopts a kind of hybrid approach by multiplying the score of BM25 and dense retrieval. To the best of our knowledge, this approach showed the state-of-the-art result on the average performance of five benchmark datasets mentioned in the next subsection.

We use \textbf{BM25}~\citep{Robertson1994-ci} and \textbf{docT5query}~\citep{Nogueira2019-hc} as models using BoW representations for queries like SPLADE-Doc. BM25 is still a strong baseline~\citep{Thakur2021-rt}. DocT5query expands documents using a generative model in addition to BM25.

Note that we did not apply domain adaptation for these baselines. We quote the results of docT5query and Cross Encoder from ~\citet{Thakur2021-rt}.

As the baseline of another UDA method, we used \textbf{GPL}~\citep{noauthor_undated-vv}, a state-of-the-art UDA method for dense retrieval. We experimented by applying GPL to SPLADE and our dense retrieval model\footnote{GPL used Margin-MSE as a loss function. The teacher model of Margin-MSE was cross-encoder/ms-marco-MiniLM-L-6-v2. When applying GPL to our dense retrieval model trained on MS MARCO, Negative examples were sampled from the top-50 results of the dense retrieval model and sentence-transformers/msmarco-MiniLM-L-6-v3. When applying GPL to SPLADE, we replaced our dense retrieval model with SPLADE.}. When we applied CAI for comparison with GPL in dense retrieval, we used weighted average pooling with IDF weights.

\subsection{Datasets and Evaluation Measures}
This study used part of BEIR ~\citep{Thakur2021-rt}. BEIR is a benchmark dataset in a zero-shot case, where no supervision data are available in the target datasets. Following the setting of BEIR, we used MS MARCO~\citep{DBLP:conf/nips/NguyenRSGTMD16} as a source domain dataset where massive supervision data are available. This means that all supervised IR models were trained using MS MARCO. We measured IR performance by nDCG@10 as BEIR. For datasets of target domains, we chose BioASK (B-ASK)~\citep{Tsatsaronis2015-bq}, NFCorpus (NFC)~\citep{Boteva2016-gs}, and TREC-COVID (T-COV)~\citep{Voorhees2021-gq} from the biomedical domain and SCIDOCS (SDOCS)~\citep{cohan-etal-2020-specter} and SciFact (SFact)~\citep{wadden-etal-2020-fact} from science domain because they have the largest vocabulary gap from the source domain. We show the vocabulary gap in Appendix~\ref{sec:voc-gao}.

We built domain-specific corpora of the biomedical and science domains for domain adaptation of an MLM. We align the domains to the target datasets. For the biomedical domain, we extracted abstracts from the latest collection of PubMed\footnote{\url{https://pubmed.ncbi.nlm.nih.gov/}}. We removed abstracts with less than 128 words from the corpus, following PubmedBERT~\citep{gu2021-pubmed}. The corpus size was approximately 17 GB. For the science domain, we used the abstracts of the S2ORC~\citep{lo-etal-2020-s2orc} corpus. We also excluded abstracts with less than 128 words from the corpus. The corpus size was approximately 7.3 GB. The resulting size of $\mathcal{V}_{final}$ was 71,694 words in the biomedical domain and 62,783 in the science domain.

\begin{table*}[!t]
    \centering
    \caption{Evaluation of our methods and other IR models by nDCG@10. The best results are in \textbf{bold}. The best results in the same category are in \textit{italics}.}
    \begin{tabular}{ l|c c c | c c | c}
    \hline
        ~ & \multicolumn{3}{|c|}{Biomedical} & \multicolumn{2}{c|}{Science} & \\ \hline
        ~ & B-ASK & NFC & T-COV & SDOCS & SFact & Ave \\ \hline \hline
        Dense & 0.377 & 0.301 & 0.716 & 0.144 & 0.571 & 0.422 \\ 
        SPLADE & 0.503 & 0.336 & 0.627 & 0.155 & 0.691 &  0.462 \\ 
        Cross Encoder & 0.523 & 0.350 & \textit{0.757} & \textit{0.166} & 0.688 & \textit{0.497} \\
        SPLADE with CAI (Ours) & \textit{0.544} & \textit{0.353} & 0.719 & 0.161 & \textit{0.708} & \textit{0.497} \\ \hline \hline
        \multicolumn{6}{l}{Bag-of-words representations of queries} \\ \hline
        BM25 & 0.515 & 0.335 & 0.581 & 0.148 & 0.674 & 0.451 \\
        docT5query & 0.431 & 0.328 & \textit{0.713} & \textit{0.162} & 0.675 & 0.462 \\
        SPLADE-Doc & 0.488 & 0.323 & 0.539 & 0.147 & 0.678 & 0.435 \\ 
        SPLADE-Doc with CAI (Ours) & \textit{0.551} & \textit{0.342} & 0.633 & \textit{0.162} & \textit{0.715} & \textit{0.480} \\
        \hline \hline
        \multicolumn{6}{l}{Hybrid with Lexical Approach} \\ \hline 
        LaPraDor & 0.511 & 0.347 & \textbf{0.779} & \textbf{0.185} & 0.697 & 0.504 \\ 
        Hybrid-SPLADE-Doc with CAI (Ours) & 0.567 & 0.347 & 0.680 & 0.162 & 0.714 & 0.494 \\
        Hybrid-SPLADE with CAI (Ours) & \textbf{0.573} & \textbf{0.357} & 0.756 & 0.165 & \textbf{0.716} & \textbf{0.514} \\ \hline\hline
        
    \end{tabular}
    \label{tab-comp-baseline}
\end{table*}

\subsection{Details of Model Training}
To train SPLADE and dense retrieval, we used Margin-MSE as a loss function\footnote{We introduce the formal expression of Margin-MSE in Appendix~\ref{sec:margin-mse}. We used  cross-encoder/ms-Marco-MiniLM-L-6-v2 as the teacher model for Margin-MSE.}.
Negative documents used in Margin-MSE were retrieved by BM25 or other retrieval methods as hard negative samples\footnote{We used negative documents distributed by sentence transformers website~\url{https://huggingface.co/datasets/sentence-transformers/msmarco-hard-negatives/resolve/main/msmarco-hard-negatives.jsonl.gz}}. The loss of SPLADE\footnote{We introduce the formal expression of SPLADE loss in Appeindix\ref{sec:loss-and-reg}} was the sum of Margin-MSE and FLOPS regularizers. The regularization weight of FLOPS for the query side $\lambda_Q$ and document side $\lambda_D$ were set as $\lambda_Q=0.08$ and $\lambda_D=0.1$, respectively, following ~\citet{Formal2021-hj}. Note that SPLADE-Doc was only used when searching, not training. We trained SPLADE and dense retrieval on one NVIDIA A100 40 GB GPU. 

For continual pretraining, we began from bert-base-uncased\footnote{https://huggingface.co/bert-base-uncased} and conducted training on eight NVIDIA A100 40 GB. We set the batch size to 32 per device and trained one epoch.

For GPL, we generated queries for each document with docT5query~\citep{Nogueira2019-hc}\footnote{Actual model is BeIR/query-gen-msmarco-t5-base-v1} using top-k and nucleus sampling (top-k: 25; top-p: 0.95). Following ~\citet{noauthor_undated-vv}, we sampled three queries per document and limited the size of the target IR corpus to 1M to reduce the computational cost when generating queries. 

We give other parameters related to training models in Appendix~\ref{sec:hyper-param}.

\section{Results}

This section compares our method with baselines. We first show the results of our approach and other IR methods. Next, we present the results of CAI and GPL as a comparison of UDA methods.

\begin{table}[!t]
    \centering
    \scriptsize
    \caption{Evaluation of our methods and GPL by nDCG@10. The best results are in \textbf{bold}. The best results in the same category are in \textit{italics}.}
    \begin{tabular}{ l|c c c | c c | c}
    \hline
        ~ & \multicolumn{3}{|c|}{Biomedical} & \multicolumn{2}{c|}{Science} & \\ \hline
        ~ & B-ASK & NFC & T-COV & SDOCS & SFact & Ave \\ \hline \hline
        \multicolumn{6}{l}{Dense Retrieval} \\ \hline
        Original & 0.377 & 0.301 & 0.716 & 0.144 & 0.571 & 0.422  \\
        GPL & \textit{0.420} & 0.325 & 0.723 & \textit{0.162} & \textit{0.654} & 0.457 \\
        CAI & 0.411 & \textit{0.329} & \textbf{0.760} & 0.148 & 0.648 & \textit{0.459} \\  \hline \hline
        \multicolumn{6}{l}{SPLADE} \\ \hline
        Original & 0.503 & 0.336 & 0.627 & 0.155 & 0.691 &  0.462 \\
        GPL & 0.513 & 0.319 & 0.708 & \textbf{0.171} & 0.676 & 0.477 \\
        CAI & \textit{0.544} & \textbf{0.353} & \textit{0.719} & 0.161 & \textit{0.708} & \textbf{0.497} \\ \hline \hline
        \multicolumn{6}{l}{SPLADE-Doc} \\ \hline 
        Original & 0.488 & 0.323 & 0.539 & 0.147 & 0.678 & 0.435 \\
        GPL & 0.491 & 0.305 & 0.562 & 0.153 & 0.649 & 0.432 \\
        CAI & \textbf{0.551} & \textit{0.342} & \textit{0.633} & \textit{0.162} & \textbf{0.715} & \textit{0.480} \\ \hline
    \end{tabular}
    \label{tab-comp-da}
\end{table}

\subsection{Comparison with other IR Methods}

Table~\ref{tab-comp-baseline}  lists the results of our method and other IR models. First, SPLADE with CAI outperformed SPLADE on all datasets. In addition, our approach showed comparable performance with Cross Encoder on the average of nDCG@10 for all datasets. These results illustrate that our method effectively fills the vocabulary and word-frequency gap for IR. Note that SPLADE can realize faster retrieval than Cross Encoder because SPLADE only has to encode queries, not concatenations of queries and documents. 

Next, SPLADE-Doc with CAI scored best on four of five datasets in other methods using BoW representations of queries. In addition, SPLADE-Doc with CAI outperformed SPLADE on all datasets. This result suggests that our approach performs quite well for BoW representations and is as fast as BM25 when searching~\footnote{We also checked the sparseness of SPLADE with CAI on the NFCorpus. The average of nonzero elements of SPLADE with CAI is 291.7, though the average document length is 175.5 with the pyserini analyzer. We consider this number to be sufficiently sparse to utilize an inverted index.}.

Finally, Hybrid-SPLADE with CAI achieved the best on the average of nDCG@10 for all datasets and outperformed LaPraDor. However, on some datasets, LaPraDor scored higher. This implies that sparse retrieval and dense retrieval learn different aspects of IR. It seems necessary in future work to research a more effective method of utilizing the complementarity of dense and sparse representations.

\subsection{Comparison of Unsupervised Domain Adaptation Methods}

Next, we compare CAI with GPL, a state-of-the-art UDA method. Table~\ref{tab-comp-da} shows the results of comparing CAI and GPL on dense retrieval, SPLADE, and SPLADE-Doc. For all IR models in Table~\ref{tab-comp-da}, our method outperformed GPL. This result shows that our method is suitable for the domain shift in vocabulary and word frequencies. Focusing on the performance difference, it was large for SPLADE and SPLADE-Doc but small for dense retrieval. This result suggests that our approach is more effective for sparse retrieval. Note that GPL deteriorates the performance of SPLADE-Doc. Our approach seems more robust for query representation in SPLADE than GPL.

\section{Ablation with AdaLM for Confirming Assumption}
\label{sec:comp-da}
We conducted ablation studies for AdaLM to confirm the assumptions presented in Section~\ref{sec:CAI}. Precisely, we considered the case of only applying vocabulary expansion or continual pretraining. In addition, we also used models trained on a domain-specific corpus from scratch.
By comparing AdaLM with continual pretraining, we confirm whether IR performance is further enhanced by expanding the vocabulary. In addition, we observe the effect of vocabulary size based on the results achieved by vocabulary expansion and the scratch models. As scratch models, we used PubmedBERT~\footnote{microsoft/BiomedNLP-PubMedBERT-base-uncased-abstract}~\citep{gu2021-pubmed} for the biomedical domain and SciBERT~\footnote{\url{https://huggingface.co/allenai/scibert_scivocab_uncased}}~\citep{beltagy-etal-2019-scibert} for the science domain.

\begin{table}[!t]
    \centering
    \small
    \caption{Ablation study using AdaLM by nDCG@10. We use SPLADE as a base model. The best results are in \textbf{bold}.}
    \begin{tabular}{ l |c| c| c}
    \hline
        ~ & Biomedical & Science & All  \\ \hline \hline
        SPLADE & 0.489 & 0.423 & 0.462 \\ \hline \hline
        \multicolumn{4}{l}{Ablation to AdaLM} \\ \hline
        Continual Pretraining & 0.509 & 0.426 & 0.476  \\ 
        Vocabulary Expansion& 0.493 & 0.416 & 0.462 \\ 
        Scratch Models & 0.000 & \textbf{0.446} & 0.178 \\ 
        AdaLM & \textbf{0.528} & 0.426  & \textbf{0.491} \\ \hline
    \end{tabular}
    \label{tab-result-ndcg}
\end{table}

Table~\ref{tab-result-ndcg} lists the result of the ablation study using SPLADE.
First, continual pretraining improved IR performance over the original SPLADE. In addition, SPLADE with AdaLM outperformed continual pretraining. These results support that vocabulary expansion enhances the effect of continual pretraining.

However, expanding vocabulary cannot improve the IR performance on average. In addition, the scratch model of the science domain outperformed AdaLM. Note that the vocabulary size of scratch is the same with the original BERT. These results show that larger dimensions themselves cannot help improve IR performance when no supervision data are available and that accurate query and document expansion is more important. 

By contrast, the scratch model of the biomedical domain failed to learn SPLADE. Thus, scratch models on the domain-specific corpus may not learn SPLADE, and AdaLM seems a more stable method than the scratch models.

We further analyzed the effect of vocabulary size on AdaLM in Section~\ref{app:vocab-size}. The result suggests that AdaLM with a larger vocabulary size tended to perform better in nDCG@10 for SPLADE.

\section{Analysis for Weight of Words}
\label{analysis-rsj-weight}
This section shows whether our method can solve the problem of exact matching of low-frequency words.

\begin{figure}[t]
 \centering
 \includegraphics[keepaspectratio, scale=0.22]{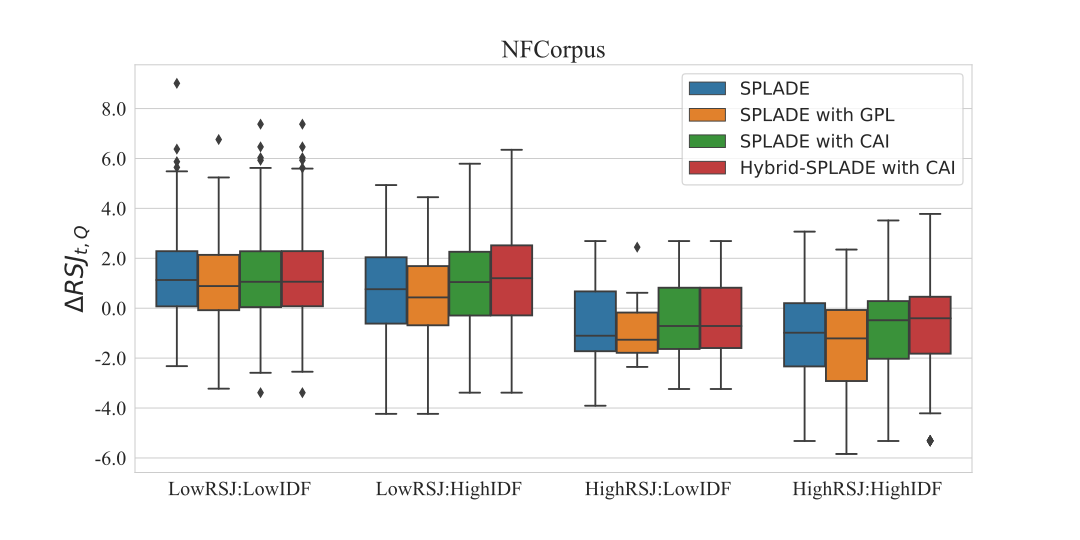}
 \caption{$\Delta \text{RSJ}_{t, Q}$ of SPLADE, SPLADE with GPL, SPLADE with CAI, and Hybrid-SPLADE with CAI.}
 \label{fig-rsj-weight-splade-nfcorpus}
\end{figure}

\citet{Formal2021-zs} analyzed IR models with an MLM, using Robertson Sp{\"a}rck Jones (RSJ) weight~\citep{UCAM-CL-TR-356}. RSJ weight measures how a token can distinguish relevant from non-relevant documents in an IR corpus. This weight also indicates an ideal weight in terms of lexical matching. We denote RSJ weight as $\text{RSJ}_{t, Q}$ for a tokenized query $Q \in \mathcal{V}^{l}$ and a token $t \in Q$.  To infer the RSJ weight of the IR models, ~\citet{Formal2021-zs} replaced relevant documents with top-K documents retrieved by the model. We express the inferred RSJ weight as $\widehat{\text{RSJ}}_{t, Q}$. We set $\text{K} = 100$, following the authors. 
We give the formal expression of $\text{RSJ}_{t, Q}$ and $\widehat{\text{RSJ}}_{t, Q}$ in Appendix~\ref{sec:rsj-weight}.

Following ~\citet{Formal2021-zs}, we take the difference between $\text{RSJ}_{t, Q}$ and $\widehat{\text{RSJ}}_{t, Q}$, i.e.,

\begin{equation}
  \Delta \text{RSJ}_{t, Q} = \text{RSJ}_{t, Q} - \widehat{\text{RSJ}}_{t, Q},
\end{equation}
as an indicator to measure the gap between the ideal RSJ weight and RSJ weight of the models.
If $\Delta \text{RSJ}_{t, Q} > 0$, an IR model overestimates the weights of tokens. Conversely, if $\Delta \text{RSJ}_{t, Q} < 0$, an IR model underestimates the weight of the tokens. 
For analysis, we also divide all tokens into $\text{HighRSJ}$ and $\text{LowRSJ}$ at the $75$-percentile. Furthermore, we split all tokens into groups of HighIDF and LowIDF at the median IDF of all tokens in queries. This analysis is conducted on the NFCorpus. The tokenizer used is the analyzer of pyserini~\footnote{https://github.com/castorini/pyserini}, which processes porter stemming and removes some stopwords.

Figure~\ref{fig-rsj-weight-splade-nfcorpus} shows the $\Delta \text{RSJ}_{t, Q}$ of SPLADE, SPLADE with GPL, SPLADE with CAI, and Hybrid SPLADE with CAI. First, SPLADE with CAI underestimates the RSJ weight less than SPLADE in the groups of HighRSJ and HighIDF. In addition, Hybrid-SPLADE with CAI is closer to $\Delta \text{RSJ}_{t, Q}=0$ than SPLADE with CAI on the same groups. This result suggests that our approach solves the problem of term matching for rare words. 
By contrast, SPLADE with GPL shows lower $\Delta \text{RSJ}_{t, Q}$ than SPLADE. GPL seems to accelerate the problem of term matching for low-frequency words. As a result, GPL may lead to lower IR performance than SPLADE, as shown in Table~\ref{tab-comp-da}.

\begin{table}[t]
    \centering
    \scriptsize
    \caption{An example of top-ranked documents for a query including a HighRSJ and HighIDF word. The example is from NFCorpus. The top-ranked document by SPLADE with CAI is correct.
    The HighRSJ and HighIDF words appearing in the query and document are labeled in \textbf{bold}.}
    \label{tab-example}
    \begin{tabular}{c|l }
    \hline
        Query & \begin{tabular}{l} \textbf{Phytate}s for the Treatment of Cancer \end{tabular} \\ \hline
        \multicolumn{2}{l}{Top-ranked documents} \\ \hline
        \begin{tabular}{c} SPLADE \\ with CAI \end{tabular} & \begin{tabular}{p{4 cm}}
        Dietary suppression of colonic cancer. Fiber or \textbf{phytate}? The incidence of colonic cancer differs widely ...
        \end{tabular} \\ \hdashline
        \begin{tabular}{c} SPLADE \end{tabular} & \begin{tabular}{p{4 cm}}
        Phytochemicals for breast cancer prevention by targeting aromatase. Aromatase is a cytochrome P450 enzyme ...
        \end{tabular} 
        \\ \hline
    \end{tabular}
\end{table}

\section{Case Study}
Finally, we confirm the case where SPLADE with CAI improves the IR performance by matching important and rare words, i.e., HighRSJ and HighIDF words. Table~\ref{tab-example} shows a pair of a query including a HighRSJ and HighIDF word and top-1 documents in NFCorpus retrieved by SPLADE with CAI and SPLADE. In the example query, phytate is a HighRSJ and HighIDF word. SPLADE with CAI ranks a correct document, including phytate, at the top. However, the top-ranked document by SPLADE does not include phytate and is incorrect. The document frequency of phytate is bottom 2\% in MS MARCO. This example supports that SPLADE with CAI successfully matches rare words in training data and can rank a correct document higher.

\section{Conclusion}
This paper presented an effective unsupervised domain adaptation method, CAI. We showed that the combination of SPLADE with CAI and the lexical approach gave a state-of-the-art performance on datasets with a large vocabulary and word-frequency gap. In addition, CAI outperformed GPL and was robust enough to show high accuracy even when BoW representations were used for query expression. Finally, our analysis showed that SPLADE with CAI addressed the problem of the exact matching of low-frequency words in training data. We believe that CAI works on smaller MLMs by distilling AdaLM because \citet{yao-etal-2021-adapt} showed that a distilled AdaLM achieved higher performance than BERT on NLP tasks and \citet{Formal2021-hj} showed that the results of SPLADE initialized with DistilBERT-base\footnote{\url{https://huggingface.co/distilbert-base-uncased}} was competitive on MS MARCO with other IR models initialized with BERT.

Integrating sparse and dense retrieval is a promising way to enhance IR performance further on out-of-domain datasets. Future work will integrate them to reveal the factors contributing to IR.

\section*{Acknowledgements}
We are grateful to the anonymous reviewers, Masahiro Kaneko, Sakae Mizuki and Ayana Niwa who commented our paper. This paper is based on results obtained from a project, JPNP18002, commissioned by the New Energy and Industrial Technology Development Organization (NEDO).

\bibliography{anthology,custom}
\bibliographystyle{acl_natbib}

\appendix
\section{Loss Function and Regulalizer}
\label{sec:loss-and-reg}

This section shows the formal expression of margin mean squared error (Margin-MSE), $\mathcal{L}_{\text{Margin-MSE}}$, and FLOPS regularizer, $\mathcal{L}^{\text{FLOPS}}$. The loss function of training SPLADE, $\mathcal{L}_{\text{SPL}}$, is 
\begin{equation}
    \mathcal{L}_{\text{SPL}} = \mathcal{L}_{\text{Margin-MSE}} + \lambda_Q \mathcal{L}^{Q}_{\text{FLOPS}} + \lambda_D \mathcal{L}^{D}_{\text{FLOPS}},
\end{equation}
where $\mathcal{L}^{Q}_{\text{FLOPS}}$ is a FLOPS regualizer of a query side and $\mathcal{L}^{D}_{\text{FLOPS}}$ is a document side

\subsection{Margin Mean Squared Error}
\label{sec:margin-mse}
 Margin-MSE~\cite{Hofstatter2020-tz} can be used to distill knowledge from Cross Encoder. Cross Encoder inferrs relevance scores by inputting concatenated queries and documents to an MLM. Now, we denote a tokenized query as $Q \in \mathcal{V}^{l}$ and tokenized document as $D \in \mathcal{V}^{m}$. $l$ and $m$ are the lengths of $Q$ and $D$, respectively. We express the concatenated text as $C_{Q, D}=[\text{CLS};Q;\text{SEP};D;\text{SEP}] \in \mathcal{V}^{m+l+3}$ and the process encoding the CLS token to a $d$-dimensinal vector as $\text{BERT}(C_{Q, D})_\text{CLS}$.  The inferred score is calculated by the dot product of the vector of the CLS token and linear layer $\mW_{\text{CLS}} \in \mathbb{R}^{d \times d}$ and a bias term of the layer $\vb_{\text{CLS}} \in \mathbb{R}^d$. Then, the score $S_{\text{CE}}(Q, D)$ is 
\begin{equation}
    S_{\text{CE}}(Q, D) = \mW_{\text{CLS}}^{\top} \text{BERT}(C_{Q, D})_{\text{CLS}} + \vb_{\text{CLS}}.
\end{equation}

Now, we assume a batch of size $B$. Then we express a query in the batch as $Q_i$, a positive document to the query as $D^+_i$, and a negative document as $D^-_i$. The difference of score $\delta_i$ between $D^+_i$ and $D^-_i$ by Cross Encoder is 
\begin{equation}
\delta_i = S_{\text{CE}}(Q_i, D^+_i) - S_{\text{CE}}(Q_i, D^-_i).
\end{equation}
Next, we express a target model for training as $M$ and a score inferred by the model between $Q$ and $D$ as $S_M(Q, D)$. The difference of scores between $D^+_i$ and $D^-_i$ by $M$ is 
\begin{equation}
    \hat{\delta_i} = S_{M}(Q_i, D^+_i) - S_{M}(Q_i, D^-_i).
\end{equation}

Finally, we can express Margin-MSE by the following equation:

\begin{equation}
    \mathcal{L}_{\text{Margin-MSE}} = \frac{1}{B}\sum_{i=1}^{B} (\delta_i - \hat{\delta_i})^2.
\end{equation}

\subsection{FLOPS Regularizer}
\label{sec:flops}
FLOPS~\citep{Paria2020Minimizing} regularizer induces sparseness to encoded vectors by neural models. We denote a query matrix as $\mQ \in \mathbb{R}^{B \times l}$, which consists of $l$-dimensional vectors of queries with batsh size $B$. In the same way, we denote a document matrix as $\mD \in \mathbb{R}^{B \times l}$. The formal expressions of FLOPS loss are  
\begin{gather}
    \mathcal{L}^Q_{\text{FLOPS}} = \sum_{j=1}^{l}(\frac{1}{B}\sum_{i=1}^B \lvert \mQ_{i,j} \rvert)^2 \\
    \mathcal{L}^D_{\text{FLOPS}} = \sum_{j=1}^{l}(\frac{1}{B}\sum_{i=1}^B \lvert \mD_{i,j} \rvert)^2.
\end{gather}

\section{Robertson Sp{\"a}rck Jones Weight}
\label{sec:rsj-weight}
\citet{Formal2021-zs} analyzed IR models with an MLM, using Robertson Sp{\"a}rck Jones (RSJ) weight~\citep{UCAM-CL-TR-356}. RSJ weight measures how a token can distinguish relevant from non-relevant documents in an IR corpus. The weight is inferred by pairs of a query $Q$ and correct documents. We denote a token of the query as $t$. Formally, the RSJ weight is 

\begin{equation}
  \text{RSJ}_{t, Q} = \log\frac{p(t|{R_Q}){p(\neg t|\neg{R_Q})}}{{p(\neg t|{R_Q})}p(t|\neg{R_Q})}.
\end{equation}

$R_Q$ is a set of relevant documents for a query $Q$. $p(t|R_Q)$ is the probability that relevant documents have token $t$. $p(t|\neg {R_Q})$ is the probability that non-relevant documents have a token $t$. Lastly, $p(\neg t|R_Q) = 1 - p(t| R_Q)$ and $p(\neg t | \neg {R_Q}) = 1 - p(t | \neg {R_Q})$.

To investigate the RSJ weight of IR models, the authors proposed the following modification:
\begin{equation}
  \widehat{\text{RSJ}}_{t, Q} = \log \frac{p(t|\hat{R}_Q^K)p(\neg t|\neg \hat{R}_Q^K)}{p(\neg t|\hat{R}_Q^K)p(t|\neg \hat{R}_Q^K)}.
\end{equation}
$\hat{R}_Q^K$ represents the top-K documents retrieved for the query $Q$ by an IR model. $p(t|\hat{R}_Q^K)$ is the probability that the top-K documents retrieved by the IR model include the token $t$. $p(t|\neg \hat{R}_Q^K)$ is the probability that top-K documents not retrieved by the IR model include the token $t$. Lastly, $p(\neg t|\hat{R}_Q^K) = 1 - p(t| \hat{R}_Q^K)$ and $p(\neg t | \neg \hat{R}_Q^K) = 1 - p(t | \neg \hat{R}_Q^K)$. 

\section{HyperParameters}

We give show hyperparameters for training the models in Tables~\ref{dense-hyper-parameter},~\ref{splade-hyper-parameter}, and~\ref{gpl-hyper-parameter}.

\label{sec:hyper-param}
\begin{table}[!t]
    \centering
    \small
    \caption{Hyper-parameters of dense retrieval}
    \label{dense-hyper-parameter}
    \begin{tabular}{l l}
    \hline
    Batch size & 64 \\
    Max document length & 300 \\
    Learning rate & 2e-5 \\
    Epoch & 30 \\
    Warmup steps & 1000 \\ \hline
    \end{tabular}
\end{table}

\begin{table}[!t]
    \centering
    \small
    \caption{Hyper-parameters of SPLADE}
    \label{splade-hyper-parameter}
    \begin{tabular}{l l}
    \hline
    Batch size & 40 \\
    Max document length & 256 \\
    Learning rate & 2e-5 \\
    Epoch & 30 \\
    Warmup steps & 1000 \\ \hline
    \end{tabular}
\end{table}

\begin{table}[!t]
    \centering
    \small
    \caption{Hyper-parameters when using GPL}
    \label{gpl-hyper-parameter}
    \begin{tabular}{l l}
    \hline
    Batch size & 24 \\
    Max document length & 350 \\
    Learning rate & 2e-5 \\
    Training steps & 140000 \\
    Warmup steps & 1000 \\ \hline
    \end{tabular}
\end{table}

\section{Vocabulary Gap from MS MARCO}
\label{sec:voc-gao}
Following ~\citet{Thakur2021-rt}, we calculated weighted Jaccard similarity $J(A, B)$ between a source dataset $A$ and target dataset $B$ in BEIR~\footnote{The target datasets were ArguAna~\citep{wachsmuth-etal-2018-retrieval}, BioASK, Climate-FEVER~\citep{leippold2020climatefever}, DBPedia-Entity~\citep{Hasibi2017-nf}, FEVER~\citep{thorne-etal-2018-fever}, FiQA~\citep{Maia2018-yf}, HotpotQA~\citep{yang-etal-2018-hotpotqa}, Natural Question~\citep{kwiatkowski-etal-2019-natural}, NFCorpus, Quora, Robust04~\citep{Voorhees2004-lw}, SCIDOCS, SciFact, TREC-COVID, and Touch{\'e}-2020~\citep{Bondarenko2020-lf}}. $J(A, B)$ is calculated by the following equation:
\begin{equation}
    J(A, B) = \frac{\sum_t \min(A_t, B_t)}{\sum_t \max(A_t, B_t)}.
\end{equation}

\begin{table}[!t]
    \centering
    \small
    \caption{Weighted Jaccard similarity between a target dataset in BEIR and MS MARCO}
    \begin{tabular}{l|r}
        \hline
        Dataset & $J(S, T)$ \\ \hline
        Natural Question & 0.523 \\
        Robust04 & 0.475 \\
        Touch{\'e}-2020 & 0.410 \\
        FiQA & 0.407 \\
        Quora & 0.395 \\
        ArguAna & 0.385 \\
        Climate-FEVER &  0.384 \\
        FEVER & 0.384 \\
        HotpotQA & 0.342 \\
        DBPedia-Entity & 0.334 \\
        SCIDOCS & 0.327 \\
        BioASK & 0.317 \\
        TREC-COVID & 0.315 \\
        NFCorpus & 0.285 \\
        SciFact & 0.273 \\ \hline
    \end{tabular}
    \label{tab-wjs}
\end{table}

Here, $A_t$ is the normalized frequency of word $t$ in a source dataset, and $B_t$ is in a target dataset. We used MS MARCO as a source dataset. Table~\ref{tab-wjs} lists the results. We can observe that the five datasets we chose for our experiment were the most dissimilar to MS MARCO.

\section{Effect of Vocabulary Size}
\label{app:vocab-size}

\begin{figure}[!t]
 \centering
 \includegraphics[keepaspectratio, scale=0.33]{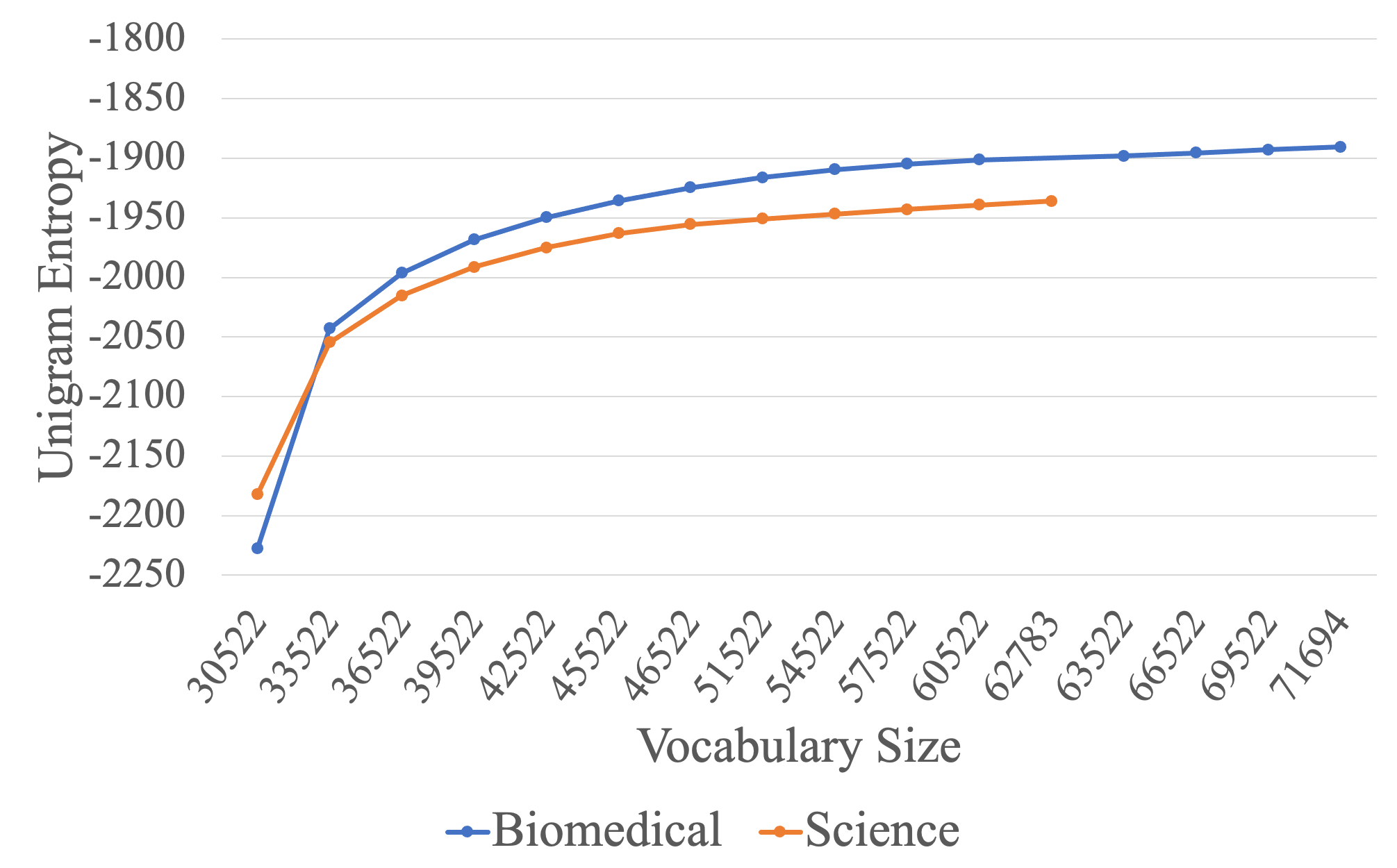}
 \caption{Unigram entropy of each vocabulary size on each domain corpus.}
 \label{fig:voc-ent}
\end{figure}

\begin{figure}[!t]
 \centering
 \includegraphics[keepaspectratio, scale=0.35]{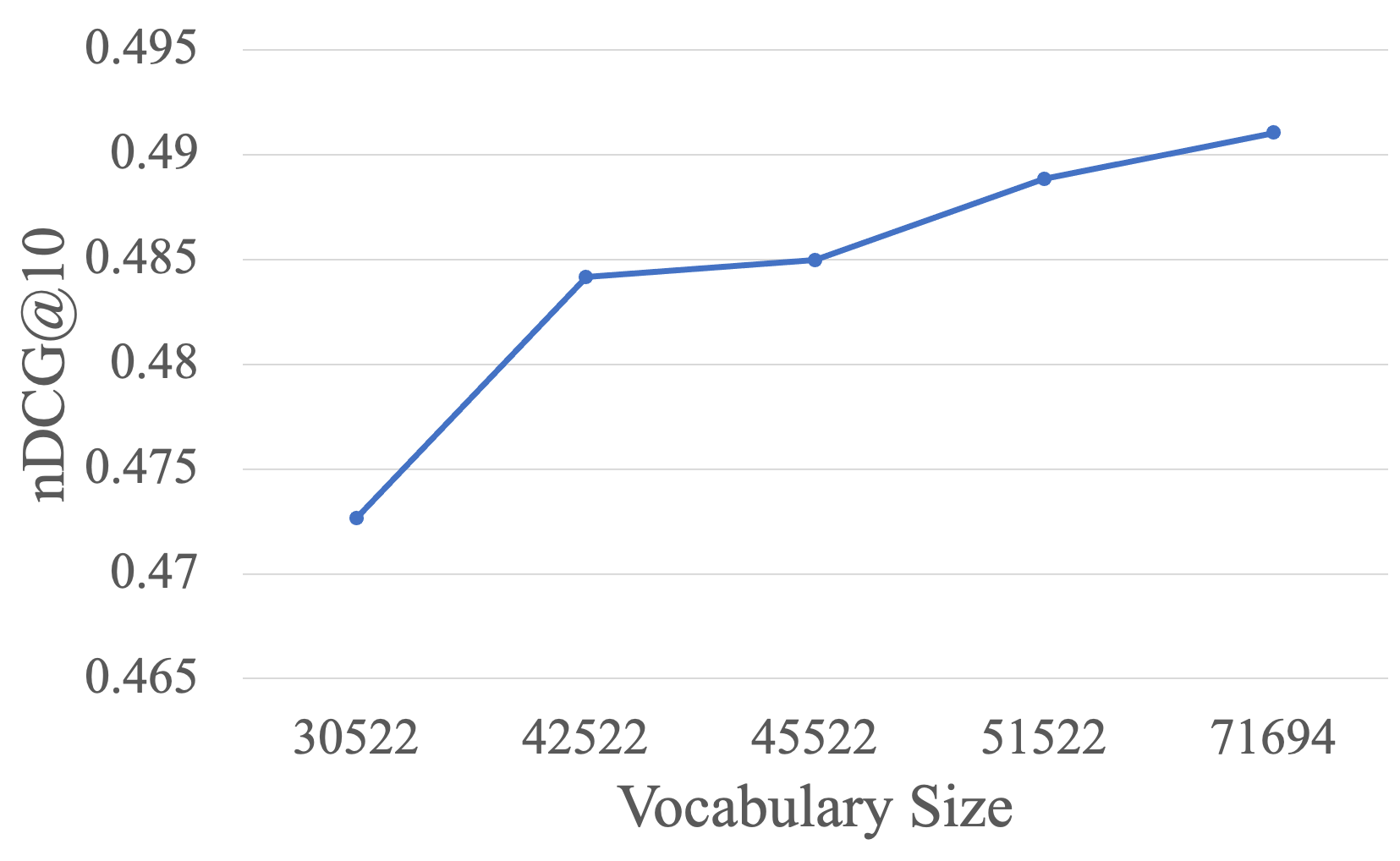}
 \caption{Performance variation with vocabulary size for SPLADE with AdaLM. Performance is measured by average nDCG@10 all datasets.}
 \label{fig:perform-vocab}
\end{figure}

To confirm the effect of vocabulary size, we experimented with the case of smaller vocabulary sizes of AdaLM. To save the computational cost, we selected several vocabulary sizes, using unigram entropy criteria $I(C)$ of MLM training corpus $C$, as by~\citet{yao-etal-2021-adapt}. 
For a tokenizer with vocabulary $\mathcal{V}$, we calculated unigram probability $p(x)$ by counting the occurrence of each sub-word $x$ in the corpus. Then, the unigram entropy $I(\vx)$ of each text sequence $\vx = (x_1, x_2, .., x_L)$ can be calculated by following equation:
\begin{equation}
    I(\vx) = \sum_{i=1}^L \log(p(x_i)).
\end{equation}
Now, we can describe the unigram entropy of the corpus $I(C)$ as   
\begin{equation}
    I(C) = \sum_{\vx \in C} I(\vx).
\end{equation}
As mentioned in Section~\ref{subsec-vocab-expand}, we increment vocabulary size from the original BERT tokenizer. We denote the vocabulary of a tokenizer in a step as $\mathcal{V}_i$ and the unigram entropy of the tokenizer as $I_i(D)$.

We prepare three additional stopping critera of vocabulary expansion vocabulary. The first is $\frac{I_i(D) - I_{i-1}(D)}{I_{i-1}(D)} < \epsilon_1$. We set $\epsilon_1 = 0.01$, as used by~\citet{yao-etal-2021-adapt}. The resulting vocabulary size was 42,522. Next, $I_i(D) - I_{i-1}(D)$  is largest in the first increment as shown in Figure~\ref{fig:voc-ent}. Thus, the next stopping criterion is  $I_i(D) - I_{i-1}(D) < \epsilon_2 (I_1(D) - I_0(D))$.  We set the coefficient $\epsilon_2 = 0.1$ and $\epsilon_2 = 0.05$. As a result, the vocabulary sizes were 45,522 and 51,522, respectively.

We present the results of SPLADE with AdaLM on the average of nDCG@10 for all datasets in Figure~\ref{fig:perform-vocab}. The figure shows the trend that the model of large vocabulary size performed better in nDCG@10. 

\section{Interaction between In-Domain Supervision Data and CAI}
We experimented in the case where in-domain supervision data were available to observe the effect of CAI with supervision data. 

We trained SPLADE and SPLADE with CAI used in our main experiment further on NFCorpus because NFCorpus has the most training pairs of a query and a relevant document in the all target datasets. The loss function for the training was MultipleNegativeRankingLoss\footnote{\url{https://www.sbert.net/docs/package_reference/losses.html\#multiplenegativesrankingloss}}~\citep{Henderson2017-mt}. Negative examples were sampled from BM25 and two dense retrieval models. One was the same with the model mentioned in Section~\ref{sec:experimental-seup}. The other was trained on NFCorpus further from the first with negative examples from BM25. We did not use Margin-MSE loss in this experiment because SPLADE models trained with Margin-MSE~\footnote{The model of cross encoder is cross-encoder/ms-marco-MiniLM-L-6-v2.} loss on NFCorpus degraded the performance. We changed $\lambda_{Q}$, $\lambda_{D}$, and batch size from the settings of Section~\ref{sec:experimental-seup}. We set the batch size at 32. We used $\lambda_{Q}=0.0006$ and $\lambda_{D}=0.0008$, following~\citet{Formal2021-hj}.

Table~\ref{tab-sup-nfcorpus} shows the results. SPLADE with supervision data of NFCorpus certainly improved nDCG@10 over the case without supervision. However, the improvement of the performance was limited. In contrast, SPLADE with CAI and supervision data showed a larger improvement. Thus, adapting MLM to the target domain is also important for SPLADE when supervision data are available.

\begin{table}[!t]
    \centering
    \small
    \caption{Experimental results with and without supervision data of NFCorpus.}
    \begin{tabular}{ l |c }
    \hline
        ~ & nDCG@10  \\ \hline \hline
        \multicolumn{2}{l}{SPLADE} \\ \hline
        Without Supervision & 0.336 \\
        With Supervision & 0.339 \\ \hline \hline
        \multicolumn{2}{l}{SPLADE with CAI} \\ \hline
        Without Supervision & 0.353 \\
        With Supervision & 0.377 \\ \hline
    \end{tabular}
    \label{tab-sup-nfcorpus}
\end{table}

\end{document}